\documentclass[conference]{IEEEtran}
\usepackage{cite}
\usepackage{amsmath,amssymb,amsfonts}
\usepackage{graphicx}
\usepackage{booktabs}
\usepackage{multirow}
\def\BibTeX{{\rm B\kern-.05em{\sc i\kern-.025em b}\kern-.08em
    T\kern-.1667em\lower.7ex\hbox{E}\kern-.125emX}}
\begin{document}

\title{Federated Binary Gating with Server-Side Vision-Language Inference for Surveillance Anomaly Classification}

\author{
\IEEEauthorblockN{
C\^{o}me-Alexis Puech\textsuperscript{1}, Sébastien Thuau\textsuperscript{1,2}, Amira Gran\textsuperscript{1}, Arthur Mennessier\textsuperscript{1}, Siba Haidar\textsuperscript{1}, Rachid Chelouah\textsuperscript{2}
}
\IEEEauthorblockA{
\textsuperscript{1}\textit{esieaLab, ESIEA}, Paris, France \quad
\textsuperscript{2}\textit{ETIS Laboratory, CNRS, UMR8051, CY Cergy Paris University}, Paris, France
}
\IEEEauthorblockA{
cpuech@et.esiea.fr, sebastien.thuau@cyu.fr, amira.gran@et.esiea.fr,\\
mennessier@et.esiea.fr, siba.haidar@esiea.fr, rachid.chelouah@cyu.fr
}
}

\maketitle

\begin{abstract}
Privacy-sensitive surveillance systems could benefit from large
vision-language models (VLMs), but such models typically require centralized
access to raw video. In federated learning settings, this challenge is
amplified by non-independent and identically distributed (non-IID) client data,
which can make direct multiclass anomaly classification unstable, especially
for rare categories. We propose a hybrid two-stage architecture that combines
a federated binary convolutional neural network (CNN) gate with server-side
zero-shot VLM inference. The lightweight LiteCNN3D gate performs local anomaly
screening and forwards only flagged videos to Qwen3-VL-8B, which assigns them
to four anomaly metaclasses. We evaluate this design on UCF-Crime grouped into
five coarse metaclasses and implement the federated stage in a real three-node
heterogeneous deployment.

In the studied setting, direct federated multiclass training collapses,
whereas the proposed decomposition yields a better trade-off between
classification quality and raw-video transmission. With fixed-threshold
routing, the federated hybrid pipeline preserves nearly the same
macro-averaged F1 score (F1-macro) as its centralized CNN+VLM counterpart while
reducing the fraction of transmitted videos to 51.4\%, although with a lower
proxy macro receiver operating characteristic area under the curve (ROC AUC)
than the centralized hybrid system. A complementary sensitivity-oriented
routing operating point increases macro ROC AUC from 0.673 to 0.692 and
reduces the false negative rate from 29.3\% to 22.9\%, but decreases F1-macro
from 0.503 to 0.485 while increasing transmission from 51.4\% to 57.9\%.
These results suggest that federation is better suited to coarse local
screening, while routing rules can be adjusted to trade server-side VLM usage
for higher anomaly sensitivity.
\end{abstract}


\section{Introduction}

Modern surveillance networks generate continuous video streams, only a small
fraction of which contains events of interest. Their analysis is constrained by
privacy, heterogeneous edge resources, and the need for semantically meaningful
anomaly recognition. Compact models such as three-dimensional convolutional
neural networks (3D CNNs) can run on edge devices, but are typically developed
in centralized settings on benchmarks such as UCF-Crime~\cite{sultani}.
Vision-language models (VLMs), by contrast, provide strong zero-shot semantic
reasoning across anomaly categories, but their computational cost makes
client-side deployment impractical.

Federated learning offers an alternative for privacy-sensitive surveillance by
keeping raw footage on-device and exchanging model updates instead~\cite{fedavg}.
However, realistic non-independent and identically distributed (non-IID) client
distributions make fine-grained multiclass anomaly classification difficult:
rare or imbalanced categories may be absent from some clients, weakening the
local training signal.

We therefore decompose the task into two stages. A lightweight binary CNN is
trained federatively and used locally to screen videos. Only videos flagged as
anomalous are transmitted to a central server, where a zero-shot VLM assigns
them to anomaly metaclasses. This design keeps coarse screening at the edge
while applying computationally expensive semantic reasoning only to a reduced
subset of inputs.

This leads to the following research question:

\begin{quote}
\textit{Can a lightweight federated binary gate, coupled with server-side
zero-shot VLM inference, approach the utility of centralized multiclass
systems while reducing raw-video transmission under non-IID data?}
\end{quote}

We study this question on UCF-Crime grouped into five coarse metaclasses. We
compare centralized CNN, federated CNN, standalone VLM, and hybrid CNN+VLM
systems in both centralized and federated configurations. In the studied
setting, direct federated multiclass learning collapses, whereas federated
binary screening remains usable. Coupled with server-side VLM inference, the
resulting hybrid pipeline preserves nearly the same F1-macro as its
centralized counterpart while substantially reducing raw-video transmission.

\noindent\textbf{Contributions.} Our contributions are as follows:
\begin{itemize}
    \item We introduce a two-stage architecture for surveillance anomaly
    classification, combining federated binary edge screening with server-side
    zero-shot VLM classification.

    \item We evaluate the system in a real three-node heterogeneous federated
    deployment, including a CPU-only client, rather than through server-side
    client emulation.

    \item We analyze the classification--transmission trade-off under
    fixed-threshold and sensitivity-oriented routing, showing that task
    decomposition remains more suitable than direct federated multiclass
    learning in the studied non-IID setting.
\end{itemize}

\section{Related Work}

\subsection{Anomaly Detection in Surveillance Video}
\label{sec:vad-related}

Sultani et al.~\cite{sultani} introduced UCF-Crime and a weakly supervised
multiple-instance ranking framework, establishing a large-scale benchmark for
real-world surveillance anomaly detection. Later methods improved temporal
anomaly scoring through multiple-instance learning, including RTFM~\cite{rtfm},
MGFN~\cite{mgfn}, and UR-DMU~\cite{urdmu}. Contrastive Language-Image
Pretraining (CLIP)-based extensions, such as CLIP-TSA~\cite{cliptsa} and
VadCLIP~\cite{vadclip}, further exploit pretrained vision-language
representations for weakly supervised anomaly discrimination.

These methods define the dominant centralized evaluation setting on UCF-Crime,
but mainly target anomaly scoring or localization rather than metaclass
classification. They also assume centralized access to surveillance footage,
which is restrictive when raw video cannot leave the capture site.

\subsection{Federated Learning for Surveillance Video Analysis}

Federated learning addresses this privacy constraint by keeping data on-device,
but its performance can degrade under non-IID client distributions. Under
Dirichlet-based heterogeneous partitioning, clients may observe disjoint or
highly imbalanced label subsets, leading to client drift and unstable
optimization~\cite{dirichlet}. FedAvg~\cite{fedavg} remains a standard
aggregation method, while FedProx~\cite{fedprox} partially mitigates this
instability.

Recent studies have extended federated learning to surveillance anomaly
analysis with multimodal models. FedVAD~\cite{fedvad} uses Generative
Pre-trained Transformer (GPT)-generated captions for semantic distillation,
while a multimodal prompt-based approach~\cite{wang2025multimodalprompt}
combines a frozen CLIP encoder with local and global prompts.


Unlike these approaches, we keep the VLM outside the federated training loop
and use it only as a server-side second stage. We also evaluate the system in
a physical heterogeneous multi-node deployment rather than through server-side
client emulation.

\subsection{Vision-Language Models for Surveillance Anomaly Recognition}

Several recent works adapt VLMs to UCF-Crime. LAVAD~\cite{lavad} uses a
frozen captioning VLM with a large language model (LLM) to infer anomaly
scores from frame captions. VERA~\cite{vera} keeps the VLM frozen and learns
guiding questions that translate anomaly detection into more explicit visual
cues. Holmes-VAU~\cite{holmesvau} fine-tunes a multimodal LLM with
hierarchical instructions and temporal sampling focused on anomaly-rich
segments. These methods improve fine-grained anomaly understanding, but assume
centralized VLM inference over the full visual input and do not address the
edge-side transmission-filtering setting considered here.

To our knowledge, prior federated video anomaly detection (VAD) work has not
examined a hybrid design in which a lightweight federated gate filters videos
before server-side VLM inference. The approach studied here targets this
setting: a federated binary model operates at the edge, while only selected
videos are transmitted for centralized zero-shot semantic classification.

\section{Methodology}

\subsection{Problem Setting and Overall Design}
\label{sec:problem_setting}

We consider privacy-sensitive surveillance anomaly classification in a
federated setting with heterogeneous clients. The objective is to reduce
raw-video transmission while preserving anomaly classification performance
under non-IID data distributions. In particular, we target a realistic
deployment setting in which lightweight video models can run locally on edge
clients, whereas large vision-language models (VLMs) are not practical to
deploy on all participating devices because of their computational and memory
cost.

To address this constraint, we decompose the task into two stages. First, a
lightweight binary classifier is trained federatively and used locally as a
gate to decide whether a video should be forwarded for further analysis.
Second, only videos flagged as anomalous are transmitted to a server-side VLM,
which assigns them to one of four anomaly metaclasses. Videos predicted as
normal remain on-device. This design reflects both the learning difficulty of
the task and the deployment constraints of the target setting: coarse anomaly
screening is kept at the edge, while fine-grained semantic reasoning is
performed centrally on a reduced subset of videos.

\subsection{Dataset and Metaclass Grouping}
\label{sec:dataset}

We use a preprocessed frame-based version of the UCF-Crime dataset~\cite{sultani},
where one frame is retained every 10 original frames to reduce storage
requirements relative to the raw videos. In this version, frames are available
at $64 \times 64$ resolution. For our experiments, these frames are resized to
$112 \times 112$ at model input to match the CNN architecture.

To improve semantic coherence and increase per-class support under federated
partitioning, we group the original UCF-Crime categories into five metaclasses:
\textit{Destruction} (Arson, Explosion), \textit{PropertyCrime} (Burglary,
Robbery, Shoplifting, Stealing, Vandalism), \textit{Violence} (Abuse, Arrest,
Assault, Fighting, Shooting), \textit{RoadAccidents}, and
\textit{NormalVideos}. For the binary routing stage, all anomalous categories
are merged into a single \textit{Anomaly} class opposed to
\textit{NormalVideos}.

Table~\ref{tab:dataset} reports the resulting video-level distribution (1,610
train, 290 test). The dataset remains substantially imbalanced:
\textit{NormalVideos} accounts for 49.7\% of training videos, whereas
\textit{Destruction} represents only 4.3\%. Under heterogeneous federated
partitioning, such minority metaclasses may become severely underrepresented or
even absent on some clients, which weakens the local training signal for
fine-grained multiclass learning.

For temperature scaling and selection of the sensitivity-oriented routing
thresholds, validation data were obtained from the original training split. The
official test set was kept untouched and used only for final reporting. All
splits were performed at the video level to avoid clip-level leakage.

\begin{table}[!t]
  \caption{Video-level metaclass distribution in the grouped UCF-Crime dataset.}
  \label{tab:dataset}
  \centering
  \begin{tabular}{lrrrr}
    \toprule
    \textbf{Metaclass} & \textbf{Train videos} & \textbf{\%} & \textbf{Test videos} & \textbf{\%}\\
    \midrule
    Destruction    &   70 &  4.3 &  30 & 10.3 \\
    PropertyCrime  &  401 & 24.9 &  49 & 16.9 \\
    Violence       &  212 & 13.2 &  38 & 13.1 \\
    RoadAccidents  &  127 &  7.9 &  23 &  7.9 \\
    NormalVideos   &  800 & 49.7 & 150 & 51.7 \\
    \midrule
    \textbf{Total} & 1{,}610 & 100 & 290 & 100 \\
    \bottomrule
  \end{tabular}
\end{table}

\subsection{LiteCNN3D Backbone}
\label{sec:cnn_architecture}

All CNN-based systems use the same LiteCNN3D backbone, with
task-specific output heads: $C=2$ for binary routing and $C=5$
for multiclass classification. Fixing the backbone ensures that
differences across CNN-based systems mainly reflect the learning
setting and task decomposition rather than model capacity.

The network has approximately 3M parameters and comprises five
3D convolutional blocks with channel sizes $\{32,64,128,256,256\}$,
each followed by batch normalization, ReLU activation, and
max-pooling. Adaptive average pooling produces a 256-dimensional
feature vector, fed to a 256--128--$C$ classification head with
dropout $p=0.3$.

\begin{figure}[!t]
  \centering
  \includegraphics[width=\columnwidth]{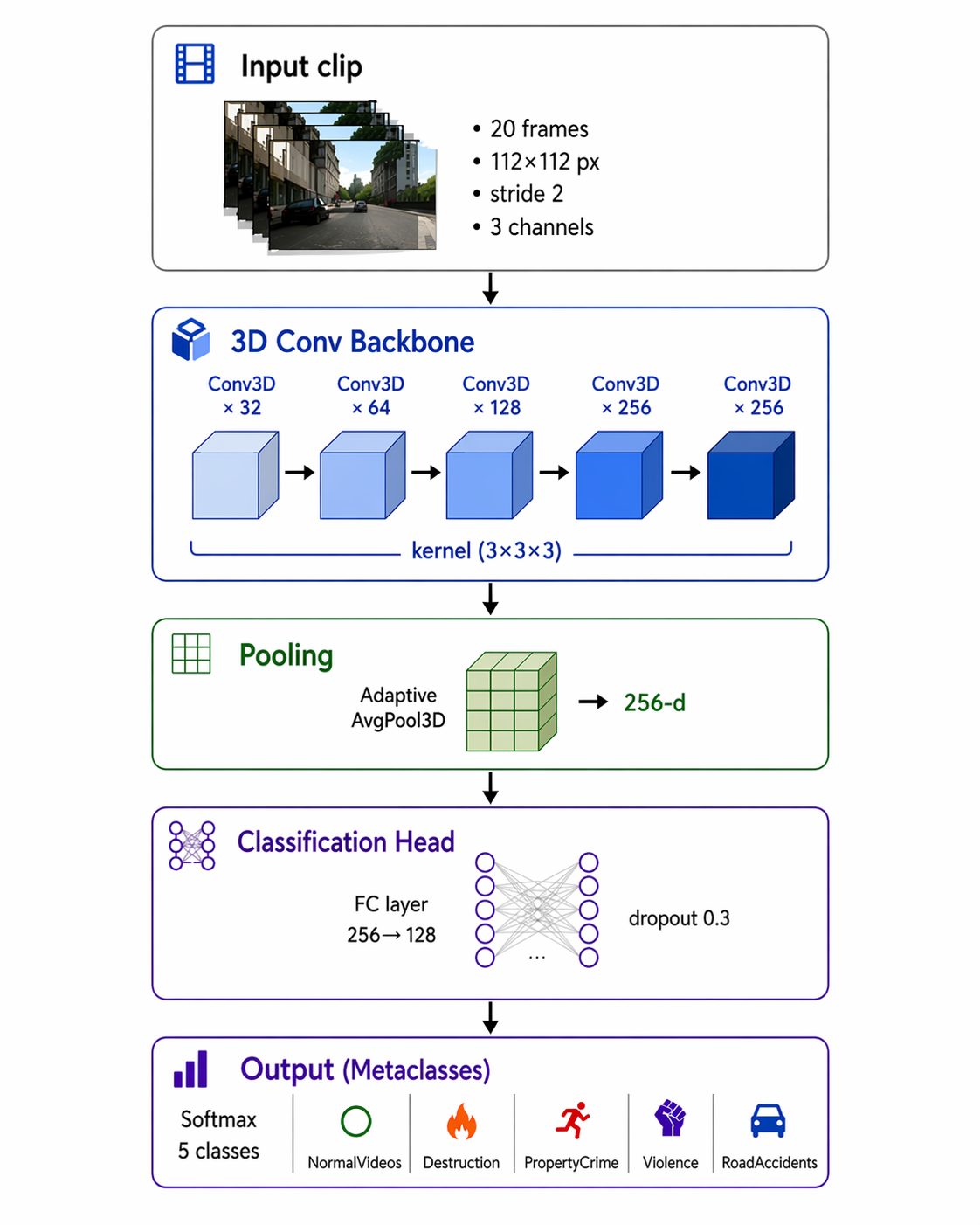}
  \caption{LiteCNN3D architecture used in the CNN-based systems. The output
  dimension of the classification head is $C=2$ for binary routing and $C=5$
  for multiclass classification.}
  \label{fig:cnn}
\end{figure}

Input clips contain 20 frames at $112 \times 112$ resolution with
temporal stride 2, sampled from the preprocessed frame sequence.
At inference time, clip-level logits are averaged over all windows
extracted from a video to obtain a video-level prediction. Fig.~\ref{fig:cnn}
summarizes the shared LiteCNN3D backbone.

\subsection{Federated Data Partitioning}
\label{sec:data_partitioning}

For the federated experiments, we use Dirichlet-based non-IID partitioning with
$\alpha = 0.30$ across 3 clients for both the binary and multiclass settings.
The difficulty of federated optimization under heterogeneous client label
distributions is well documented~\cite{dirichlet}. Using the same $\alpha$ in
both settings ensures that the comparison between federated binary routing and
federated multiclass learning is not confounded by a different degree of data
heterogeneity.

Partitioning is performed at the video level to preserve temporal
coherence. Since the reported results rely on a single Dirichlet draw, they
should be interpreted as a case study of one heterogeneous partition rather
than as an estimate averaged over multiple sampled partitions.

\subsection{Two-Stage Hybrid Pipeline}
\label{sec:hybrid_pipeline}

The hybrid pipeline operationalizes the decomposition introduced in
Section~\ref{sec:problem_setting}. A federated binary LiteCNN3D gate first
filters videos locally. Videos predicted as normal remain on-device, whereas
flagged videos are transmitted to the server and classified by Qwen3-VL-8B into
one of the four anomaly metaclasses. This design reflects both the difficulty
of learning fine-grained multiclass boundaries under federation and the
practical constraint that large VLMs are more naturally deployed server-side
than on heterogeneous edge clients.

\subsubsection{Stage 1 --- Federated Binary CNN Gate}

A binary LiteCNN3D model is trained federatively using binary
cross-entropy with logits and positive-class weighting to mitigate class
imbalance. At inference time, clip-level logits are averaged into a video-level
logit, whose sigmoid score is thresholded at $\tau=0.7$ for the fixed-threshold
routing baseline.

We use a threshold above the default decision value of 0.5 to define a
conservative communication-oriented operating point. Since the gate acts as a
transmission filter rather than a final anomaly classifier, increasing $\tau$
reduces unnecessary server forwarding and prioritizes communication and
inference cost reduction over maximum anomaly recall.

\subsubsection{Stage 2 --- Server-Side Vision-Language Model}

Videos flagged by the binary gate are transmitted to the server and analyzed by
Qwen3-VL-8B~\cite{qwen3} in a zero-shot setting. The VLM receives 24 uniformly
sampled frames in temporal order and assigns each video to one of four anomaly
metaclasses: \textit{Destruction}, \textit{PropertyCrime}, \textit{Violence},
or \textit{RoadAccidents}. Videos rejected by the binary gate are directly
labeled as \textit{NormalVideos} and are not transmitted.

The structured prompt used for the VLM is:

\begin{quote}
\textit{``You are analyzing surveillance camera footage that has already been
flagged as containing a criminal incident. Your task: identify which ONE
category best describes the crime shown in these frames. Categories:
Destruction (visible fire, flames or smoke from deliberate arson, or a sudden
explosion); PropertyCrime (breaking into a building, theft, vandalism,
robbery); Violence (assault, fistfight, abuse, shooting, arrest);
RoadAccidents (vehicle collision or traffic accident). Rules: reply with ONLY
the category name, nothing else, no explanation or punctuation.''}
\end{quote}

\begin{figure}[!t]
  \centering
  \includegraphics[width=\columnwidth]{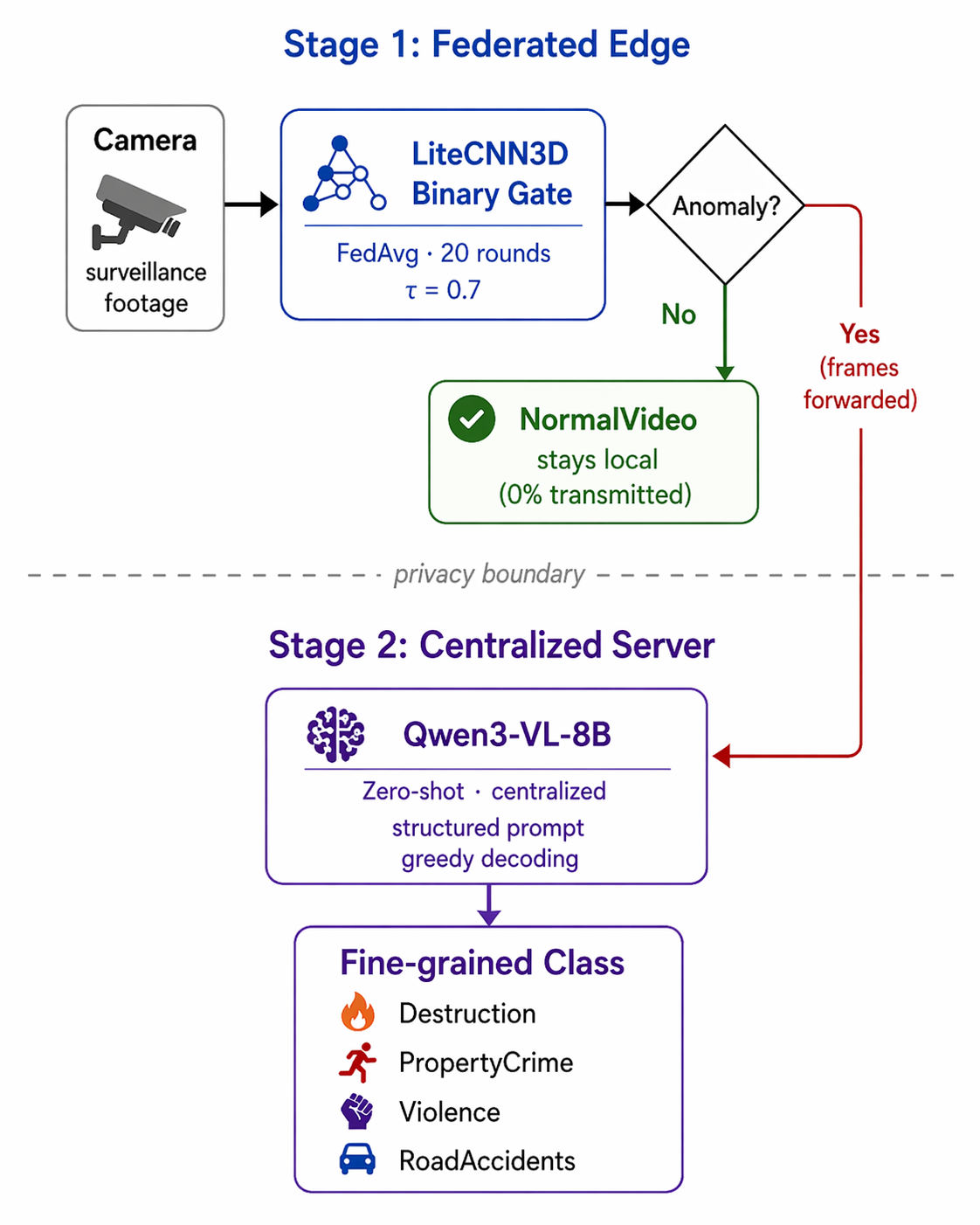}
  \caption{Two-stage hybrid pipeline: a federated binary CNN gate runs locally
  on each client, and only videos flagged as anomalous are forwarded to the
  server for zero-shot VLM classification.}
  \label{fig:pipeline}
\end{figure}

Inference uses greedy decoding with sampling disabled and generation limited to
20 new tokens. The output is parsed into one of the four anomaly metaclasses.
Figure~\ref{fig:pipeline} summarizes the proposed two-stage pipeline and makes
explicit that only videos accepted by the federated gate cross the privacy
boundary for server-side semantic classification.

\section{Experiments and Results}

\subsection{Compared Systems}
\label{sec:compared_systems}

We compare seven systems designed to isolate the roles of federation, task
granularity, and server-side semantic reasoning: (1) a centralized binary CNN,
(2) a federated binary CNN, (3) a centralized multiclass CNN, (4) an
end-to-end federated multiclass CNN, (5) a standalone server-side VLM, and
(6--7) two hybrid CNN+VLM pipelines, one centralized and one federated. The
federated hybrid pipeline is the proposed method, while the centralized hybrid
pipeline serves as its non-federated counterpart.

\textbf{Centralized Binary CNN.} A binary LiteCNN3D model is trained on the
pooled training set for \textit{Anomaly} vs.\ \textit{NormalVideos}
classification. It provides a centralized reference for the routing stage and
helps isolate the effect of federation on binary screening.

\textbf{Federated Binary CNN.} The same binary LiteCNN3D model is trained
federatively and evaluated as the routing stage of the proposed hybrid
pipeline. Evaluating it independently allows us to measure whether binary
screening remains viable under non-IID client data.

\textbf{Centralized Multiclass CNN.} A multiclass LiteCNN3D model is trained
on the pooled training set for five-class prediction, providing the centralized
counterpart to federated multiclass learning and a reference for direct
CNN-based metaclass classification without task decomposition.

\textbf{Federated Multiclass CNN.} The multiclass LiteCNN3D model is trained
end-to-end in the federated setting to test whether direct multiclass learning
remains viable under non-IID client data. This baseline represents the direct
federated alternative to the proposed hybrid decomposition.

\textbf{Standalone Zero-shot VLM.} Qwen3-VL-8B is applied server-side to all
290 test videos, without binary gating, as a fully centralized semantic
classification reference. It estimates the behavior of the VLM in isolation,
without any transmission filtering.

\textbf{Centralized Hybrid CNN+VLM.} A centralized binary gate is followed by
the same server-side VLM, isolating the effect of federation on the routing
stage while preserving the hybrid decomposition.

\textbf{Federated Hybrid CNN+VLM.} This is the proposed method: a federated
binary gate filters videos locally, and only flagged videos are forwarded to
the server-side VLM for anomaly metaclass classification.

\subsection{Evaluation Protocol}
\label{sec:evaluation_protocol}

\textbf{Binary gate evaluation.} The centralized and federated binary CNN gates
are evaluated at the video level. Clip logits are averaged over all clips from
a video before applying the sigmoid. Metrics include accuracy, F1-binary,
receiver operating characteristic area under the curve (ROC AUC), expected
calibration error (ECE; lower is better), false positive rate (FPR), and false
negative rate (FNR).

\textbf{Multiclass and end-to-end evaluation.} All multiclass and hybrid
pipeline results are reported at the video level. For CNN-based multiclass
models, clip logits are averaged over all clips of a video before applying the
softmax. Metrics include macro ROC AUC (one-vs-rest), per-class AUC and
recall, F1-macro, and the fraction of videos transmitted to the server. Unless
stated otherwise, results use the final available checkpoint, without test-set
checkpoint selection.

\textbf{Validation-based routing selection.} The official test set is used
only for final evaluation. Temperature scaling and routing-threshold selection
for the sensitivity-oriented routing rule are performed using validation data
from the original training split. This prevents the sensitivity-oriented
operating point from being selected directly on the test set.

\textbf{Hybrid pipeline AUC construction.} The binary gate outputs an anomaly
score $p \in [0,1]$. Since the VLM returns hard labels rather than calibrated
probabilities, ROC AUC is computed from proxy 5-class score vectors derived
from $p$ and the final predicted label. The \textit{NormalVideos} score is set
to $1-p$. If the final prediction is an anomaly metaclass, that class receives
score $p$ and the remaining anomaly classes receive 0; if the final prediction
is \textit{NormalVideos}, the anomaly mass $p$ is distributed uniformly across
the four anomaly classes. Macro ROC AUC is computed in a one-vs-rest setting.
These values should be interpreted as ranking proxies, not calibrated class
probabilities.

\textbf{Standalone VLM evaluation.} The standalone VLM uses a 5-class prompt
containing the four anomaly metaclasses and \textit{NormalVideos}. The prompt
explicitly favors anomaly predictions unless normality is clear, with
\textit{NormalVideos} treated as a fallback class. Since the model does not
provide continuous class scores, ROC AUC is computed from one-hot encoded hard
predictions. Its AUC should therefore be viewed as only approximately
comparable to the AUC values obtained from softmax-based CNN outputs or proxy
hybrid scores.

\subsection{Federated Training Setup}
\label{sec:federated_setup}

Federated training is implemented with Flower~\cite{flower} in a real
three-node deployment on heterogeneous hardware: an NVIDIA RTX 3080 (10\,GB
VRAM), an NVIDIA RTX 4060 (8\,GB VRAM), and a laptop equipped with an AMD
Ryzen 7 7840U and integrated Radeon 780M graphics, used for CPU-only training.
Flower's synchronous protocol is used, so the server waits for the slowest
client before starting the next communication round.

Model aggregation is performed with FedAvg over 20 communication rounds, with
2 local epochs per round. Optimization uses Adam with cosine annealing
($\eta_0 = 3 \times 10^{-4}$, $\eta_{\min} = 10^{-6}$). The batch size is set
to 4 in order to accommodate the CPU-only client. For multiclass training, we
use cross-entropy loss with label smoothing ($\epsilon = 0.05$) and
inverse-frequency class weighting. For the binary gate, we use a binary
classification objective with positive-class weighting.

\subsection{Binary Routing Performance}
\label{sec:binary_results}

We first compare the centralized and federated binary gates at the video level,
using the final available checkpoint without test-set checkpoint selection.
Table~\ref{tab:binary_final_comparison} shows that the federated binary gate
achieves slightly stronger hard-decision performance, reaching 0.709 accuracy
and 0.729 F1-binary, compared with 0.625 and 0.693 for the centralized
counterpart.

However, the centralized model retains a higher ROC AUC (0.790 vs.\ 0.766) and
better calibration (ECE 0.103 vs.\ 0.224). This suggests that federation
degrades score reliability, although the final routing decisions remain
competitive. For the proposed hybrid pipeline, this trade-off is acceptable
because the first-stage model acts primarily as a transmission filter rather
than as a calibrated probability estimator.

\begin{table}[!t]
  \caption{Video-level performance of the binary routing gate in centralized and federated settings. Acc. = accuracy; F1-bin. = binary F1 score; ROC AUC = receiver operating characteristic area under the curve; ECE = expected calibration error.}
  \label{tab:binary_final_comparison}
  \centering
  \begin{tabular}{lcccc}
    \toprule
    \textbf{Setting} & \textbf{Acc.} & \textbf{F1-bin.} & \textbf{ROC AUC} & \textbf{ECE} \\
    \midrule
    Centralized & 0.625 & 0.693 & 0.790 & 0.103 \\
    Federated   & 0.709 & 0.729 & 0.766 & 0.224 \\
    \bottomrule
  \end{tabular}
\end{table}

\subsection{Multiclass and Hybrid System Comparison}
\label{sec:multiclass_results}


Table~\ref{tab:global_comparison} compares the systems at the global video
level. The standalone VLM obtains the highest macro ROC AUC, although this
value should be interpreted cautiously because it is derived from one-hot hard
predictions rather than continuous scores. The centralized hybrid pipeline
achieves the best F1-macro, indicating that binary gating followed by semantic
VLM classification is more effective than centralized multiclass CNN
classification alone. The proposed federated hybrid pipeline preserves nearly
the same F1-macro as its centralized hybrid counterpart (0.503 vs.\ 0.506),
while reducing the fraction of transmitted videos from 79.7\% to 51.4\%.

In contrast, direct federated multiclass learning collapses in this setting:
F-CNN-Multi reaches only 0.048 F1-macro and yields the weakest overall
performance. This is the main empirical result of the paper: under the studied
non-IID conditions, federated learning remains usable for coarse binary
screening, but not for direct end-to-end multiclass anomaly classification.

\begin{table}[!t]
  \caption{Global video-level comparison of multiclass systems. C-CNN: centralized multiclass LiteCNN3D; C-VLM: standalone server-side Qwen3-VL-8B; C-CNN+VLM: centralized hybrid pipeline; F-CNN+VLM: proposed federated hybrid pipeline; F-CNN-Multi: end-to-end federated multiclass LiteCNN3D. Sent (\%) denotes the fraction of test videos forwarded to the server.}
  \label{tab:global_comparison}
  \centering
  \scriptsize
  \setlength{\tabcolsep}{3pt}
  \renewcommand{\arraystretch}{1.05}
  \resizebox{\columnwidth}{!}{%
  \begin{tabular}{lccccc}
    \toprule
    \textbf{Metric} & \textbf{C-CNN} & \textbf{C-VLM} & \textbf{C-CNN+VLM} & \textbf{F-CNN+VLM} & \textbf{F-CNN-Multi} \\
    \midrule
    Macro AUC      & 0.783 & 0.848 & 0.757 & 0.673 & 0.650 \\
    F1-macro       & 0.420 & 0.472 & \textbf{0.506} & 0.503 & 0.048 \\
    Sent (\%)      & 100\% & 100\% & 79.7\% & 51.4\% & 0\% \\
    \bottomrule
  \end{tabular}%
  }
\end{table}

Table~\ref{tab:classwise_comparison} further highlights the difference between
ranking-oriented and assignment-oriented behavior. The standalone VLM achieves
the highest AUC on most classes, whereas the centralized hybrid pipeline
obtains the strongest recall on most anomaly classes, including
\textit{PropertyCrime} (0.918), \textit{RoadAccidents} (0.739), and
\textit{Violence} (0.395). This is consistent with its superior F1-macro.

The proposed federated hybrid pipeline remains below the centralized hybrid,
but retains meaningful class-wise utility, with competitive recall for
\textit{NormalVideos} (0.667), \textit{PropertyCrime} (0.633), and
\textit{RoadAccidents} (0.652). By comparison, F-CNN-Multi behaves almost as a
single-class predictor, with recall 1.000 on \textit{Destruction} but 0.000 on
\textit{NormalVideos}, \textit{PropertyCrime}, and \textit{RoadAccidents}, and
only 0.026 on \textit{Violence}. These class-wise results reinforce the main
conclusion: hybrid decomposition preserves usable behavior under non-IID data,
whereas direct federated multiclass learning does not.

\begin{table}[!t]
  \caption{Per-class video-level AUC and recall for the evaluated multiclass systems. C-CNN: centralized multiclass LiteCNN3D; C-VLM: standalone server-side Qwen3-VL-8B; C-CNN+VLM: centralized hybrid pipeline; F-CNN+VLM: proposed federated hybrid pipeline; F-CNN-Multi: end-to-end federated multiclass LiteCNN3D. Bold values highlight the most relevant comparisons discussed in the text.}
  \label{tab:classwise_comparison}
  \centering
  \scriptsize
  \setlength{\tabcolsep}{3.2pt}
  \renewcommand{\arraystretch}{1.05}
  \resizebox{\columnwidth}{!}{%
  \begin{tabular}{llccccc}
    \toprule
    \textbf{Metric} & \textbf{Class} & \textbf{C-CNN} & \textbf{C-VLM} & \textbf{C-CNN+VLM} & \textbf{F-CNN+VLM} & \textbf{F-CNN-Multi} \\
    \midrule
    \multirow{5}{*}{\textbf{AUC}}
      & Destruction    & 0.733 & \textbf{0.828} & 0.688 & 0.579 & 0.587 \\
      & NormalVideos   & 0.870 & \textbf{0.889} & 0.770 & 0.687 & 0.834 \\
      & PropertyCrime  & 0.790 & 0.834 & \textbf{0.835} & 0.708 & 0.694 \\
      & RoadAccidents  & 0.804 & \textbf{0.900} & 0.884 & \textbf{0.792} & 0.534 \\
      & Violence       & 0.719 & \textbf{0.788} & 0.606 & 0.600 & 0.600 \\
    \midrule
    \multirow{5}{*}{\textbf{Recall}}
      & Destruction    & 0.200 & 0.333 & \textbf{0.467} & 0.167 & 1.000 \\
      & NormalVideos   & 0.680 & \textbf{0.953} & 0.347 & \textbf{0.667} & 0.000 \\
      & PropertyCrime  & 0.714 & 0.286 & \textbf{0.918} & \textbf{0.633} & 0.000 \\
      & RoadAccidents  & 0.522 & 0.478 & \textbf{0.739} & \textbf{0.652} & 0.000 \\
      & Violence       & 0.158 & 0.132 & \textbf{0.395} & \textbf{0.263} & 0.026 \\
    \bottomrule
  \end{tabular}%
  }
\end{table}

\subsection{Sensitivity-Oriented Routing for Hybrid Inference}
\label{sec:routing_analysis}

We further analyze a second routing operating point aimed at increasing anomaly
sensitivity. The rule combines the temperature-calibrated CNN anomaly score
with predictive entropy, but it is not intended to isolate the independent
contribution of the entropy term. Rather, it provides an exploratory
sensitivity-oriented operating point for studying how the hybrid pipeline
behaves when more uncertain or borderline videos are forwarded to the
server-side VLM.

A video is kept locally as \textit{NormalVideos} only when both the calibrated
anomaly score and the predictive entropy fall below selected thresholds;
otherwise, it is forwarded to the server-side VLM. Temperature scaling and
threshold selection for this sensitivity-oriented rule were performed using
validation data from the original training split. The selected configuration is
$T=1.2$, $p_{\text{low}}=0.5$, and $h_{\text{low}}=0.69$. Since the selected
entropy threshold is close to the maximum binary entropy, this operating point
should be interpreted primarily as a more permissive, sensitivity-oriented
routing configuration rather than as evidence that entropy alone improves
routing.

As reported in Table~\ref{tab:routing_comparison}, this operating point shifts
the behavior of the hybrid pipeline. Compared with fixed-threshold routing, it
increases macro AUC from 0.673 to 0.692 and reduces the false negative rate
from 29.3\% to 22.9\%, meaning that fewer anomalous videos are missed at the
anomaly-vs-normal level. However, this comes at the cost of higher VLM usage,
with transmission increasing from 51.4\% to 57.9\%, and a lower F1-macro on
the test set, decreasing from 0.503 to 0.485.

Class-wise recall further explains this trade-off. The sensitivity-oriented
rule improves recall for several anomaly metaclasses, including
\textit{RoadAccidents} (65.2\% to 73.9\%), \textit{Destruction} (16.7\% to
26.7\%), and \textit{Violence} (26.3\% to 31.6\%), while keeping
\textit{PropertyCrime} nearly unchanged. This improvement in anomaly recall is
obtained at the expense of lower \textit{NormalVideos} recall (66.7\% to
60.0\%), which is consistent with the higher transmission rate.

Overall, these results highlight the role of routing as a controllable
mechanism rather than a uniformly better replacement for fixed-threshold
routing. The sensitivity-oriented rule forwards more videos to the server-side
VLM and reduces missed anomalies, whereas the fixed-threshold rule remains
more favorable in terms of F1-macro and transmission cost.

\begin{table}[!t]
  \caption{Fixed-threshold and sensitivity-oriented routing in the federated hybrid pipeline. Sent (\%) denotes the fraction of test videos forwarded to the server.}
  \label{tab:routing_comparison}
  \centering
  \scriptsize
  \setlength{\tabcolsep}{4pt}
  \renewcommand{\arraystretch}{1.05}
  \begin{tabular}{lcc}
    \toprule
    \textbf{Metric} & \textbf{Fixed threshold} & \textbf{Sensitivity-oriented} \\
    \midrule
    Macro AUC   & 0.673 & 0.692 \\
    F1-macro    & 0.503 & 0.485 \\
    Sent (\%)   & 51.4\% & 57.9\% \\
    FNR (\%)    & 29.3\% & 22.9\% \\
    \bottomrule
  \end{tabular}
\end{table}

\section{Discussion}
\label{sec:discussion}

The results show that task decomposition is more suitable than direct
federated multiclass learning in the studied setting. Under the same non-IID
partitioning and training conditions, the end-to-end federated multiclass CNN
collapsed, whereas the federated binary gate remained operational. This
suggests that the binary anomaly-vs-normal decision is more robust to
heterogeneous client distributions than fine-grained multiclass classification.

This behavior is consistent with the difficulty of the two tasks. Binary
screening requires a coarser decision boundary, while multiclass classification
must separate imbalanced anomaly categories that may be unevenly represented or
absent across clients. As a result, client drift and minority-class
under-representation affect direct federated multiclass learning more strongly.

The hybrid design also matches practical deployment constraints. Lightweight
CNN inference can run on heterogeneous edge clients, including CPU-only
devices, whereas running a large VLM on every client would often be
impractical. Centralizing the VLM allows semantic reasoning to be applied only
to videos selected by the local gate, reducing raw-video transmission while
keeping the expensive model outside the federated loop.

The routing results should be interpreted as operating points rather than
universal deployment rates. Fixed-threshold routing forwards 51.4\% of the test
videos, while sensitivity-oriented routing increases transmission to 57.9\%.
This shift improves macro AUC and reduces the false negative rate, but lowers
F1-macro on the test set. The sensitivity-oriented rule should therefore be
interpreted as an operating point that prioritizes anomaly sensitivity rather
than as a uniformly better routing strategy. This highlights a key property of
the proposed pipeline: routing acts as a controllable mechanism to balance
server-side VLM usage and anomaly detection sensitivity. In this context,
stricter filtering reduces transmission, whereas sensitivity-oriented routing
reduces missed anomalies.

Since UCF-Crime contains more anomalous videos than typical surveillance
streams, which are usually dominated by normal footage, both transmission
rates may overestimate those observed in real deployments.

The centralized baselines should be interpreted as upper references rather than
as target operating points. The objective here is not to surpass centralized
processing, but to identify useful operating points under privacy,
communication, and edge-resource constraints. From that perspective, the
federated hybrid design remains effective despite a residual gap to
centralized performance.

The conclusions remain bounded by the experimental scope. The study relies on a
single Dirichlet partition, heuristic routing rules, and FedAvg under short
local training. In addition, the sensitivity-oriented routing rule was selected
as a single operating point and was not subjected to a dedicated ablation
isolating the contribution of its entropy component. Its thresholds were also
selected using validation data drawn from the original training split, which
may yield optimistic threshold choices. The study does not characterize
variability across partitions, routing configurations, or federated optimizers.
Even so, the results provide meaningful empirical evidence because they are
obtained in a real federated deployment rather than through server-side client
simulation, and because the compared settings are matched to the design
question under study.

Finally, metric interpretation requires caution. Macro AUC values are not
strictly comparable across systems because CNNs provide continuous scores,
whereas the VLM and hybrid systems rely on hard labels or proxy score vectors.
The strongest evidence is therefore the joint observation that the federated
multiclass baseline collapses, while the federated hybrid pipeline maintains
usable F1-macro and class-wise behavior, with routing rules allowing different
sensitivity--transmission trade-offs. Overall, the contribution should be
interpreted as a deployment-oriented result rather than a state-of-the-art
classification claim.

\section{Conclusion}

This paper investigated whether a lightweight federated binary gate combined
with server-side zero-shot VLM inference can approach the utility of
centralized multiclass systems while reducing raw-video transmission under
non-IID data. In the studied setting, the proposed federated hybrid pipeline
achieved an F1-macro close to its centralized hybrid counterpart while
forwarding only 51.4\% of the test videos to the server under fixed-threshold
routing.

We further showed that a sensitivity-oriented routing operating point changes
the behavior of the hybrid system: it increases macro ROC AUC and reduces the
false negative rate, but lowers F1-macro and increases raw-video transmission.
This highlights the role of routing as a controllable mechanism to balance VLM
usage and anomaly sensitivity.

These results support the use of federation for coarse local anomaly screening
rather than direct fine-grained semantic classification. The proposed design
therefore provides a practical compromise for privacy-sensitive surveillance:
raw videos are filtered locally by a lightweight CNN, and only selected videos
are transmitted for centralized VLM-based semantic classification.

This study remains limited to a single Dirichlet partition, heuristic routing
rules, and FedAvg-based optimization in a three-node heterogeneous deployment.
Future work should evaluate multiple partition draws, explore adaptive routing
strategies, improve gate calibration, and investigate stronger or more
efficient multimodal back-ends.


\end{document}